\documentclass[11pt,a4paper]{article}
\usepackage[hyperref]{naaclhlt2018}
\usepackage{times}
\usepackage{latexsym}
\usepackage{graphicx}
\usepackage{subcaption}

\usepackage{CJKutf8}

\usepackage{tikz}
\usepackage{tikz-qtree}
\usepackage{tikz-qtree-compat}
\usepackage{tikz-dependency}

\usepackage{url}
\usepackage{textcomp}

\aclfinalcopy %

\title{Detecting Syntactic Features of Translated Chinese}

\author{Hai Hu, Wen Li, Sandra K\"ubler  \\
Indiana University \\
{\tt \{huhai,wl9,skuebler\}@indiana.edu} }

\date{}

\begin{document}
\maketitle
\begin{abstract}
\begin{CJK}{UTF8}{gbsn}
We present a machine learning approach to distinguish texts translated to Chinese (by humans) from texts originally written in Chinese, with a focus on a wide range of syntactic features. Using Support Vector Machines (SVMs) as classifier on a genre-balanced corpus in translation studies of Chinese, we find that constituent parse trees and dependency triples as features without lexical information perform very well on the task, with an F-measure above 90\%, close to the results of lexical \textit{n}-gram features, without the risk of learning topic information rather than translation features.
Thus, we claim syntactic features alone can accurately distinguish translated  from original Chinese. 
Translated Chinese exhibits an increased use of determiners, subject position pronouns, NP + ``的'' as NP modifiers, multiple NPs or VPs conjoined by ``、'', among other structures. 
We also interpret the syntactic features with reference to previous translation studies in Chinese, particularly the usage of pronouns.
\end{CJK}
\end{abstract}

\section{Introduction}

Work in translation studies has shown that translated texts differ significantly in subtle and not so subtle ways from original, non-translated texts. For example, \citet{volansky2013features} show that the prefix \textit{mono-} is more frequent in Greek-to-English translations because epistemologically it originates from Greek. Also, the structure of modal verb, infinitive, and past participle (e.g. \textit{must be taken}) is more prevalent in translated English from 10 source languages.

We also know that a machine learning based approach can distinguish translated from original texts with high accuracy for Indo-European languages such as Italian \citep{baroni2005new}, Spanish \citep{ilisei2010identification}, and English \citep{volansky2013features,
CLpaper2012lgModelsforMT,koppel2011ACLtranslationese}. Features used in those studies include common bag-of-words features, such as word \textit{n}-grams, as well as part-of-speech (POS) \textit{n}-grams, function words, etc. Although such surface features yield very high accuracy (in the high nineties), they do not contain much deeper syntactic information, which is key in interpreting textual styles. Furthermore, despite the large amount of research on Indo-European languages, few studies have quantitatively investigated either lexical or syntactic features of translated Chinese, and to our knowledge, no automatic classification experiments have been conducted for this language. 

Thus the purpose of this paper is two-fold: First, we perform translated vs.\ original text classification on a balanced corpus of Chinese, in order to verify whether translationese in Chinese is as real as it is in Indo-European languages, and to discover which structures are prominent in translated but not original Chinese texts. Second, we show that using only syntactic features without any lexical information, such as context-free grammar (CFG), subtrees of constituent parses, and dependency triples, perform almost as well as lexical \textit{n}-gram features, confirming the translationese hypothesis from a purely syntactic point of view. These features are also easily interpretable for linguists interested in syntactic styles of translated Chinese. We analyze the top syntactic features ranked by a common feature selection algorithm, and interpret them with reference to previous studies on translationese features in Chinese.

\section{Related Work}\label{sec:related}

\subsection{Translated vs.\ Original Classification\label{sec:transVSori:litreview}}
The pioneering work of \citet{baroni2005new} is one of the first to use machine learning methods to distinguish translated and original (Italian) texts. They experimented with word/lemma/POS \textit{n}-grams and mixed representations and reached an F-measure of 86\% using recall maximizing combinations of SVM classifiers. In the mixed \textit{n}-gram representation, they used inflected wordforms for function words, but replaced content words with their POS tags. The high F-measure (85.2\%) with such features shows that ``function word distributions and shallow syntactic patterns'' without any lexical information can already account for much of the characteristics of translated text.

\citet{volansky2013features} is a very comprehensive study that investigated translationese in English by looking at original and translated English from 10 source languages, in a European parliament corpus. While they mainly aimed to test translational universals, e.g. simplification, explicitation, etc., the classification accuracy with SVMs using features such as POS trigrams (98\%), function words (96\%),  function word \textit{n}-grams (100\%) provided more evidence that function words and surface syntactic structures may be enough for the identification of translated text. 

For Chinese, however, there are very few quantitative studies on translationese \citep[apart from][etc.]{xiao2015corpus,hu2010}. \citet{xiao2015corpus} built a comparable corpus containing 500 original and translated Chinese texts respectively, from four genres. They used statistical tests (log-likelihood tests) to find statistical differences between translated and original Chinese with regard to the frequency of mostly lexical features.  They discovered, for example, that translated text use significantly more pronouns than the original texts, across all genres. But they were unable to investigate the syntactic contexts in which those overused pronouns occur most often. 

For them, syntactic features were examined through word \textit{n}-grams, similar to previous studies on Indo-European languages, but no text classification task was carried out. 

\subsection{Syntactic Features in Text Classification}

Although \textit{n}-gram features are more prevalent in text-classification tasks, deep syntactic features have been found useful as well.
In the Native Language Identification (NLI) literature, which in many respects is similar to the task of detecting translations,  various forms of context-free grammar (CFG) rules are often used as features \citep{bykh2014syntaxFeatNLI,syntaxFeatNLI2011Wong}. \citet{bykh2014syntaxFeatNLI} showed that using a form of normalized counts of \textit{lexicalized} CFG rules plus \textit{n}-grams as features in an ensemble model performed better than all other previous systems. \citet{syntaxFeatNLI2011Wong} reported that using unlexicalized CFG rules (except for function words) from two parsers yielded statistically higher accuracy than simple lexical features (function words, character and POS \textit{n}-grams). 

Other approaches have used rules of tree substitution grammar (TSG) \citep{explicitImplSynFeat-post2013,TSGinNLI2012Swanson} in NLI. \citet{TSGinNLI2012Swanson} compared the results of CFG rules and two variants of TSG rules 
and showed that TSG rules obtained through Bayesian methods reached the best results. 

Nevertheless, such deep syntactic features are rarely used, if at all, in the identification of translated texts. This is the gap that we hope to fill.

\begin{table}[t]
\centering
\scalebox{0.90}{
\begin{tabular}{llllll} \hline
	\# texts & news & \begin{tabular}[c]{@{}l@{}}general\\ prose\end{tabular} & science & fiction & total \\ \hline
	LCMC & 88 & 206 & 80 & 111 & 485 \\
	ZCTC & 88 & 206 & 80 & 111 & 485 \\ \hline
\end{tabular}
}
\caption{Distribution of texts across genres}
\label{table:corpus}
\end{table}

\section{Experimental Setup} 

\subsection{Dataset}
We use the comparable corpus by \citet{xiao2015corpus}, which is composed of 500 original Chinese texts from the Lancaster Corpus of Modern Chinese (LCMC), and another 500 human translated Chinese texts from the Zhejiang-University Corpus of Translated Chinese (ZCTC). All texts are of similar lengths (\texttildelow2000 words), and from different genres. There are four broad genres: news, general prose, science, and fiction (see Table~\ref{table:corpus}), and 15 second-level categories. We exclude texts from the second-level categories ``science fiction'' and ``humor'' (both under fiction) since they only have 6 and 9 texts respectively, which is not enough for a classification task. 

LCMC \cite{LCMC} was originally designed for ``synchronic studies of Chinese and the contrastive studies of Chinese and English'' \cite[see][chapter 4.2]{xiao2015corpus}. It includes written Chinese sampled from 1989 to 1993, amounting to about one million words. ZCTC was created specifically for translation studies ``as a comparable counterpart of translated Chinese'' to LCMC  \cite[][pp. 48]{xiao2015corpus}, with the same genre distribution and also one million words in total. The texts in ZCTC are sampled in 2001, 
all translated by human translators, with 99\% originally written in English (pp. 50). 

Both corpora contain texts that are segmented and POS tagged, processed by the corpus developers using the 2008 version of ICTCLAS \cite{ICTCLAS}, a common tagger used in Chinese NLP research. 
However, only the segmentation is used in this study since our parser uses a different  POS tagset.%

In this study, we perform 5-fold cross validation on the whole dataset and then evaluate on the full set of 970 texts.

\subsection{Pre-Processing and Parser}
\begin{CJK}{UTF8}{gbsn}
We remove URLs and headlines, normalize irregular ellipsis (e.g. ``。。。'', ``....'') to ``……'', change all half-width punctuations to full-width, so that our text is compatible with the Chinese Penn Treebank \cite{xue2005CTB}, which is the training data for the \texttt{Stanford CoreNLP} parser \cite{corenlpmanning2014} used in our study. 

\end{CJK}

\subsection{Features\label{features}}
Character and word \textit{n}-gram features can be considered upper bound and baseline. On the one hand, they have been used extensively (see Section~\ref{sec:related}), but on the other hand, they partially encode topic information rather than stylistic differences because of their lexical nature.   Consequently, while they are very informative in the current setup, they may not be useful if we want to use the trained model on other texts.

For syntactic features, we use various forms of constituent and dependency parses of the sentences. %
We extract the following features based on either type of parse using the \texttt{CoreNLP} parser with its pre-trained parsing model.

\begin{figure}[t]
\centering
\scalebox{0.8}{
\begin{tikzpicture}
\begin{CJK}{UTF8}{gbsn}
\tikzset{every tree node/.style={align=center,anchor=north}}
\Tree [.ROOT [.IP [.NP [.PN 我们\\\textit{we} ] ] [.VP [.ADVP [.AD 一起\\\textit{together} ]] [.VP [.VV 照\\\textit{take} ] [.NP [.QP [.CLP [.M 幅\\\textit{CL} ]]] [.NP [.NN 像\\\textit{picture} ] ] ] ] ] [.PU 。\\\textit{.} ] ] ] 
\end{CJK}
\end{tikzpicture}
}
\caption{Example constituent tree of the Chinese sentence meaning \textit{We take a picture together}\label{tree}}
\end{figure}
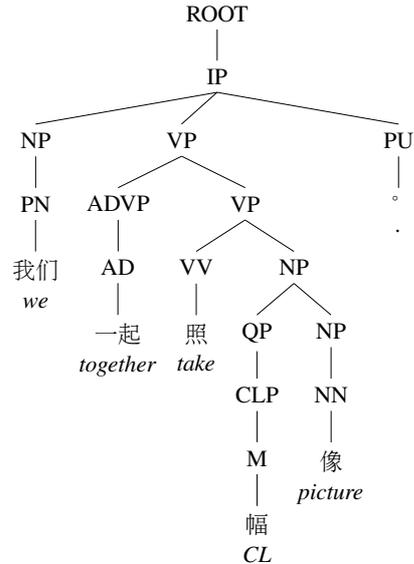
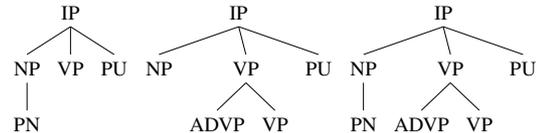
\begin{figure}[t]
\centering
\scalebox{0.7}{
\begin{tikzpicture}
\Tree [.IP [.NP [.PN ] ] [.VP ] [.PU ] ]
\end{tikzpicture}

\centering
\begin{tikzpicture}
\Tree [.IP [.NP ] [.VP [.ADVP ] [.VP ] ] [.PU ] ] 
\end{tikzpicture}

\centering
\begin{tikzpicture}
\Tree [.IP [.NP [.PN ]] [.VP [.ADVP ] [.VP ] ] [.PU ] ] 
\end{tikzpicture}
}
\caption{All subtrees of depth 2 with root IP in the tree from  Figure~\ref{tree} \label{tree2}}
\end{figure}

\subsubsection{Context-Free Grammar} 

\paragraph{Context-free grammar rules} (CFGR) We use the count of each CFG rule extracted from the parse trees.

\paragraph{Subtrees} Subtrees are defined as any part of the constituent tree of any depth, closely following the data-oriented parsing (DOP) paradigm \citep{dop2003,goodman1998parsing}. Our features differ from the DOP model as well as TSG \citep{bayesianTSG2009,parsingTSG2011,TSGinNLI2012Swanson} in that we do not include any lexical information in order to exclude topical influence from content words. Thus no lexical rules are considered, and POS tags are considered to be the leaf nodes (Figure~\ref{tree2}).

We experiment with subtrees of depth up to 3 since the number of subtrees grows exponentially as the depth increases. With depth 3, we are already facing more than 1 billion features. Performing subtree extraction and feature selection becomes difficult and time consuming. Also note that CFGRs are essentially subtrees of depth 1. So with increasing maximum depth of subtrees, we  test fewer local relations in constituent parses. In the future, we plan to use Bayesian methods \citep{bayesianTSG2009} to sample from all the subtrees.

We also conduct separate experiments using subtrees headed by a specific label (we only look at NP, VP, IP, and CP, since they are the most frequent types of subtrees). For example, using NP subtrees as features will inform us how important the noun phrase structure is in identifying translationese.

\subsubsection{Dependency Graphs}

Dependency relations, as well as the head and dependent are extracted to construct the following features.

\begin{CJK}{UTF8}{gbsn}
\paragraph{depTriple} We combine the POS of a head and its dependent along with the dependency relation, e.g., [VV, nsubj, PN] describes a dependency relation of a nominal subject (nsubj) between a verb (VV) and a pronoun (PN).

\paragraph{depPOS} Here only the POS tags of the head and dependent are used, e.g., [VV, PN].

\paragraph{depLabel} Only the dependency relation, e.g., [nsubj].

\begin{figure}[t]
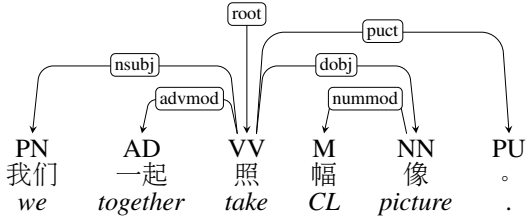

\centering
\scalebox{0.9}{
	\begin{dependency}
		\begin{deptext}[column sep=0.4cm]
			PN \& AD \& VV \& M \& NN \& PU \\
			我们 \& 一起 \& 照 \& 幅 \& 像 \& 。 \\
			\textit{we} \& \textit{together} \& \textit{take} \& \textit{CL} \& \textit{picture} \& . \\
		\end{deptext}
		\depedge{3}{1}{nsubj}
		\depedge{3}{2}{advmod}
		\depedge{5}{4}{nummod}
		\depedge{3}{5}{dobj}
		\depedge{3}{6}{puct}
		\deproot{3}{root}
	\end{dependency}
}
\caption{Example dependency graph \label{depGraphEg}}
\end{figure}

\paragraph{depTripleFuncLex} Same as depTriple, except when the word is a function word, we use the lexical item instead of the POS. e.g. [VV, nsubj, 我们] where 
``我们" (\textit{we}) is a function word (Figure~\ref{depGraphEg}).
\end{CJK}

It should be noted that no lexical information are included in our syntactic features, except for the function words in depTripleFuncLex.

\subsubsection{Combination of Features}
If combined feature sets work significantly better than one feature set alone, we can draw the conclusion that they model different characteristics of translationese. We experiment with combination of CFGR/subtree and depTriple features.

\begin{CJK}{UTF8}{gbsn}
\end{CJK}

\subsection{Classifier and Feature Selection}
For the machine learning experiments, we use support vector machines, in the implementation of the svm.SVC classifier in scikit-learn \citep{scikit-learn}.  We perform 5-fold cross validation and average over the results. When extracting the folds, we perform stratified sampling across genres so that both training  and test data are balanced. Since the number of CFGR/subtree features is much greater than the number of training texts, we perform feature selection by filtering using information gain \citep{IGliu2016,syntaxFeatNLI2011Wong} to choose the most discriminative features. Information gain has been shown to select highly discriminative, frequent features for similar tasks \citep{liu:kuebler:ea:14}. 
We experiment with different numbers of features, ranging between the values of 100, 1~000, 10~000, and 50~000.

\section{Results}

\subsection{Empirical Evaluation}

\begin{CJK}{UTF8}{gbsn}
First we report the results based on lexical and POS features in Table~\ref{table:upperbound} (F-measure). 

\paragraph{Character \textit{n}-grams} perform the best, achieving an F-measure of 95.3\%, followed by word \textit{n}-grams with an F-measure of 94.3\%. Both settings include content words that indicate the source language. 
In fact, out of the top 30 character \textit{n}-gram features that predict translations, 4 are punctuations, e.g., the first and family name delimiter ``·" in the translations of English names and parentheses ``（）"; 11 are function words, e.g. ``的'' (particle), ``可能'' (\textit{maybe}), ``在'' (\textit{in/at}), and many pronouns (\textit{he, I, it, she, they}); all others are content words, where ``斯'' (\textit{s}) and ``尔'' (\textit{r}) are at the very top, mainly because they are common transliterations of foreign names involving ``s'' and ``r'', followed by ``公司'' (\textit{company}), ``美国'' (\textit{US}), ``英国'' (\textit{UK}), etc. Lexical features have been extensively analyzed in \citet{xiao2015corpus}, and they reveal little concerning syntactic styles of translated text; thus we will refrain from analyzing them here.

\paragraph{POS \textit{n}-grams} also produce good results (F-measure of 93.9\%), confirming previous research on Indo-European languages \citep{baroni2005new,koppel2011ACLtranslationese}. Since they are not lexicalized and thus avoid a topical bias, they provide a better comparison to syntactic features.
\end{CJK}

\begin{table}[t]
\centering
\begin{tabular}{ll}
\hline
Features & F-measure (\%) \\ \hline
char \textit{n}-grams(1-3) & 95.3 \\
word \textit{n}-grams(1-3) & 94.3 \\
POS \textit{n}-grams(1-3) & 93.9 \\ \hline
\end{tabular}
\caption{Results for the lexical and POS features}
\label{table:upperbound}
\end{table}

\begin{table}[t]
\centering
\scalebox{0.9}{
\begin{tabular}{ll}
	\hline
	Features & F (\%)\\ \hline 
	\textit{Unlexicalized syntactic features} & \\ \hline
	CFGR & 90.2 \\
	subtrees: depth 2 & 90.9 \\
	subtrees: depth 3 & 92.2 \\ \hline
	depTriple & 91.2 \\
	depPOS &  89.9 \\
	depLabel & 89.5 \\ 
	depTripleFuncLex & 93.8 \\ \hline 
	\textit{Combinations of syntactic features} & \\ \hline
	CFGR + depTriple & 90.5 \\
	subtree\_d2 + depTriple & 91.0 \\ \hline
	\textit{POS \textit{n}-grams + unlex syn features} & \\ \hline
	POS + subtree\_d2 & 93.6 \\
	POS + depTriple & 93.4 \\
	POS + subtree\_d2 + depTriple & 93.8 \\ \hline 
	\textit{Char \textit{n}-grams + unlex syn features} & \\ \hline
	char + subtree + depTriple & 94.4 \\
	char + pos + subtree + depTriple & 95.5 \\ \hline %
\end{tabular}
}
\caption{Classification based on syntactic features\label{table:results}}
\end{table}

\paragraph{Syntactic features:} Table~\ref{table:results} presents the result for the syntactic features described in Section~\ref{features}. The best performing unlexicalized syntactic features can reliably classify texts into ``original'' and ``translated'', with F-measures greater than 90\%, which are close to the performance of the purely lexicalized features in Table~\ref{table:upperbound}. 
This suggests that although lexical features do achieve slightly better results, syntactic features alone can capture most of the differences between original and translated texts.

Note that when we increase the depth of constituent parses from  1 (CFGR) to subtrees of depth 3, the F-measure increases by 2 percent, which is a highly significant difference (McNemar \citep{mcnemar1947note} on the 0.001 level). Thus, including deeper constituent information proves helpful in detecting the syntactic styles of texts.  

However, combination of different types of syntactic features does not increase the accuracy over the dependency results. 
Adding syntactic features to POS \textit{n}-gram or character \textit{n}-gram features decreases the POS \textit{n}-gram results slightly, thus indicating that both types of features cover the same information, and POS \textit{n}-grams are a good approximation of shallow syntax.
The lack of improvement when adding syntactic features may also be attributed to their unlexicalized nature in this study. Our syntactic features are completely unlexicalized, whereas research in NLI has shown that CFGR features need to include at least the function words to give higher accuracy \citep{syntaxFeatNLI2011Wong}. 
Although this suggests that in terms of classification accuracy, unlexicalized syntactic features cannot provide more information than \textit{n}-gram features, we can still draw some very interesting observations about styles of translated and original texts, many of which are not possible with simple \textit{n}-gram features. We will discuss those in the following sections.

\begin{table}[t]
\centering
\scalebox{0.9}{
\begin{tabular}{ll}
	\hline
	Features & F (\%) \\ \hline 
	CFGR NP  & 86.4 \\
	CFGR VP  & 85.6 \\
	CFGR IP  & 86.6 \\
	CFGR CP  & 68.4 \\ \hline
	subtrees NP d2 & 86.0 \\
	subtrees VP d2 & 85.6 \\
	subtrees IP d2 & 89.0 \\
	subtrees CP d2 & 71.6 \\ \hline
	subtrees NP d3 & 83.6 \\
	subtrees VP d3 & 86.7 \\
	subtrees IP d3 & 86.9 \\
	subtrees CP d3 & 77.7 \\ \hline
\end{tabular}
}
\caption{Results for individual subtrees}\label{table:resSubtree}
\end{table}

\begin{CJK}{UTF8}{gbsn}
\subsection{Constituency Features}
The top ranking CFG features are shown in Table~\ref{tab:CFGfeats}. The top three features in translated section (bottom half) of the table tell us that pronouns (PN) and determiners (DT) are indicative of translated text. We will discuss pronouns in Section~\ref{sec:pronoun}; as for determiners, dependency graph features in Table~\ref{tab:depfeats} further show that among them, ``该'' (\textit{this}), ``这些'' (\textit{these}) and ``那些'' (\textit{those}) are the most prominent. The parenthesis rule (PRN) captures another common feature of translation, i.e., giving the original English form of proper nouns (``加州大学洛杉矶分校（UCLA）'') or putting translator's notes in parentheses. Furthermore, the prominence of the two rules \texttt{NP $\rightarrow$ DNP NP} and \texttt{DNP $\rightarrow$ NP DEG} in translation indicates that when an NP is modified by another NP, translators tend to add the particle ``的'' (DE; DEG for DE Genitive) between the two NPs, for example:

\begin{itemize}
\itemsep0em
\item (NP (DNP (NP 美国) (DEG 的)) (NP 政治)). Gloss: ``US DE politics'', i.e. US politics
\item (NP (DNP (NP 舆论) (DEG 的)) (NP 谴责)). Gloss: ``media DE criticism'', i.e. criticism from the media
\item (NP (DNP (NP 脑) (DEG 的)) (NP 供血)). Gloss: ``brain DE blood supply'', i.e. cerebral circulation
\end{itemize}

In all three cases above, ``的'' can be dropped, and the phrases remain grammatical. But there are many cases where ``的'' is mandatory in the ``NP modifying NP'' structure. Thus, it is easier to use ``的'', since it is almost always grammatical, but decisions when to drop ``的'' are much more subtle. Translators seem to make the safer decision by always using the particle after the NP modifiers, thus making the structure more frequent.

\begin{table}[t]
\centering
\scalebox{0.90}{
	\begin{tabular}{lll}
		Rank & CFGR & Predicts \\ \hline
		2.0 & VP $\rightarrow$ VP PU VP & original \\
		5.0 & VP $\rightarrow$ VP PU VP PU VP & original \\
		10.0 & NP $\rightarrow$ NN & original \\
		10.2 & NP $\rightarrow$ NN PU NN & original \\
		13.6 & IP $\rightarrow$ NP PU VP & original \\
		14.8 & NP $\rightarrow$ NN NN & original \\
		15 & NP $\rightarrow$ ADJP NP & original \\
		16.6 & IP $\rightarrow$ NP PU VP PU & original \\
		18.2 & VP $\rightarrow$ VV & original \\
		19.6 & VP $\rightarrow$ VV NP & original \\ \hline

		1.0 & NP $\rightarrow$ PN & translated \\
		4.0 & NP $\rightarrow$ DP NP & translated \\
		6.2 & DP $\rightarrow$ DT & translated \\
		6.6 & IP $\rightarrow$ NP VP PU & translated \\
		6.8 & PRN $\rightarrow$ PU NP PU & translated \\
		6.8 & NP $\rightarrow$ NR & translated \\
		10.0 & CP $\rightarrow$ ADVP IP & translated \\
		10.6 & NP $\rightarrow$ DNP NP & translated \\
		16.4 & ADVP $\rightarrow$ CS & translated \\
		16.8 & DNP $\rightarrow$ NP DEG & translated \\  \hline

	\end{tabular}
}
\caption{Top 20 CFGR features; rank averaged across 5-fold CV}
\label{tab:CFGfeats}
\end{table}

Now we turn to features of subtrees rooted in specific syntactic categories. The classification results are shown in Table~\ref{table:resSubtree}. Using only NP-headed rules gives us an F-measure of 86.4\%. Larger subtrees fare slightly worse, probably indicating data sparsity. However, these results mean that noun phrases alone often provide enough information whether the text is translated. 

Table~\ref{tab:NPfeats} shows the top 20 CFGR features headed by an NP. This gives us an idea of the distinctive structures of noun phrases in original and translated texts. Apart from the obvious over-use of pronouns (PN) and determiner phrases (DP) for NPs in translated text, there are other very interesting patterns: For original Chinese, nouns inside a complex noun phrase tend to be conjoined by a Chinese specific punctuation ``、"(similar to the comma in ``I like apples, oranges, bananas, etc.''), indicated by the high ranking of NP rules involving PU. This punctuation is most often used to separate elements in a list, and a check using \texttt{Tregex} \citep{tregex2006} for the parsed sentences retrieves many phrases like the following from the LCMC corpus: ``全院医生、护士最先挖掘的..." (\textit{doctors, nurses from the hospital first dug out...}). In contrast, in translated Chinese, those nouns are more likely to be conjoined by a conjunction (CC), exemplified by the following example from the ZCTC corpus: ``对经济\textbf{和}股市非常敏感" (\textit{very sensitive to the economy \textbf{and} the stock market.}). Here, to conjoin doctors and nurses, or the economy and the stock market, either ``、" or ``\textit{and}'' is grammatical, but original texts favor the former while the translated text, probably influenced by English, prefers the conjunction.

\begin{table}[t]
\centering
\scalebox{0.90}{
	\begin{tabular}{rll}
		Rank & NP CFGR & Predicts \\ \hline
		2.0 & NP $\rightarrow$ NN & original \\
		4.0 & NP $\rightarrow$ NN NN & original \\
		5.4 & NP $\rightarrow$ NN \textbf{PU} NN & original \\
		6.2 & NP $\rightarrow$ ADJP NP & original \\
		9.8 & NP $\rightarrow$ NN \textbf{PU} NN \textbf{PU} NN & original \\
		9.8 & NP $\rightarrow$ NP ADJP NP & original \\
		12.2 & NP $\rightarrow$ NP PU NP & original \\
		12.6 & NP $\rightarrow$ NN NN NN & original \\
		14.6 & NP $\rightarrow$ NP NP & original \\
		17.0 & NP $\rightarrow$ NP QP NP & original \\
		18.4 & NP $\rightarrow$ QP NP & original \\ \hline
		1.0 & NP $\rightarrow$ PN & translated \\
		4.2 & NP $\rightarrow$ DP NP & translated \\
		6.0 & NP $\rightarrow$ NR & translated \\
		7.2 & NP $\rightarrow$ DNP NP & translated \\
		14.4 & NP $\rightarrow$ QP DNP NP & translated \\
		16.2 & NP $\rightarrow$ NP PRN & translated \\
		16.2 & NP $\rightarrow$ NR \textbf{CC} NR & translated \\
		18.2 & NP $\rightarrow$ NP \textbf{CC} NP & translated \\ \hline
	\end{tabular}
}
\caption{Top 20 NP features (PN: pronoun; NR: proper N; CC: coordinating conjunction)
}
\label{tab:NPfeats}
\end{table}

\subsection{Dependency Features}
Features based on dependency parses have similar F-measures, but should be easier to obtain than subtrees of depth greater than 1. Using the lexical items for function words (depTripleFuncLex) can further improve the results, showing that the choice of function words is indeed very indicative of translationese. A selection of top ranking depTripleFuncLex features is shown in Table~\ref{tab:depfeats}. 

Chinese-specific punctuations such as ``、'' predicts original Chinese text, as we have already seen, but notice that it is also often used to conjoin verbs (VV\_PUNCT\_、).
Translated texts, in contrast, use more determiners (\textit{these}, \textit{such}, \textit{those}, \textit{each}, etc.) and pronouns (\textit{he}, \textit{they}, etc.), which  will be discussed in more detail in the following section. These results are in accordance with previous research on translationese in Chinese \citep{HeYang2008,xiao2015corpus}.

\begin{table}[t]
\centering
\scalebox{0.80}{
	\begin{tabular}{rllr} \hline
		Rank & Feature    & Predicts & Gloss 
		\\ \hline
		1.0  & VV\_CONJ\_VV  & original %
		\\
		2.4 & VV\_PUNCT\_，  & original %
		\\
		2.6 & NN\_PUNCT\_、  & original %
		\\
		4.8 & VV\_PUNCT\_、  & original %
		\\
		11.0 & NN\_CONJ\_NN  & original %
		\\
		18.0 & NN\_DET\_各   & original & each 
		\\
		21.4 & VA\_PUNCT\_，  & original %
		\\
		25.0 & NN\_ETC\_等   & original & etc. 
		\\
		28.2 & VV\_PUNCT\_：  & original %
		\\
		33.2 & VV\_PUNCT\_！  & original %
		\\
		39.0 & NN\_DEP\_三   & original & three 
		\\
		41.2 & NN\_DET\_全   & original & all 
		\\
		42.6 & VA\_\textbf{NSUBJ}\_NN  & original %
		\\
		77.2 & VV\_DOBJ\_NN & original %
		\\
		94.8 & VV\_\textbf{NSUBJ}\_NN & original %
		\\ \hline
		5.4 & VV\_\textbf{NSUBJ}\_我  & translated & I \\

		8.2 & VV\_ADVMOD\_将  & translated & will \\
		10.0 & VV\_\textbf{NSUBJ}\_他  & translated & he \\
		10.2 & NN\_DET\_该   & translated & this \\
		11.6 & NN\_DET\_这些  & translated & these \\
		14.0 & NR\_CASE\_的  & translated & DE \\
		17.0 & VV\_\textbf{NSUBJ}\_他们  & translated& they \\
		24.0 & VV\_\textbf{NSUBJ}\_她  & translated & she \\
		27.6 & 他\_CASE\_的   & translated & his \\
		29.6 & NN\_NMOD:ASSMOD\_他 & translated & he \\
		31.0 & VV\_PUNCT\_。  & translated & period \\
		33.6 & VV\_ADVMOD\_但是  & translated & but \\
		35.6 & VV\_\textbf{NSUBJ}\_你  & translated & you \\
		35.8 & VV\_ADVMOD\_如果  & translated & if \\
		37.6 & VV\_MARK\_的  & translated & DE \\
		37.8 & NN\_DET\_任何  & translated & any \\
		40.6 & VV\_CASE\_因为  & translated & because \\
		41.2 & NR\_CC\_和   & translated & and \\
		44 & NN\_DET\_那些  & translated & those \\
		47.2 & VV\_NSUBJ\_它  & translated & it \\ 
		191.0 & VV\_DOBJ\_它 & translated & it \\ \hline
	\end{tabular}
}
\caption{Top depTripleFuncLex features}
\label{tab:depfeats}
\end{table}

\end{CJK}

\section{Analyzing Features: Pronouns} \label{sec:pronoun}

In this section, we discuss one example where syntactic features provide unique information about the stylistic differences between original and translated Chinese that cannot be extracted from lexical sequences, yielding new insights into translationese in Chinese: We have a closer look at the use of pronouns. 
For this investigation, we examine the top 100 subtrees with depth 2, selected by information gain.

Our results not only confirm the previous finding that pronoun usage is more prominent in translated Chinese \citep[among others, see Section~\ref{sec:transVSori:litreview}]{HeYang2008,xiao2015corpus}, but also provide more insights on the details of pronoun usage in translated Chinese, by looking at the syntactic structures that involve a pronoun (PN) and their ranking after applying the feature ranking algorithm (see Table~\ref{pronounFeat}).

\begin{table}[t]
\centering
\scalebox{0.80}{
\begin{tabular}{rlll}
	\hline
	Rank & Feature & Function \\ \hline
	1.0 & (NP \textbf{PN}) & NA \\
	2.2 & (IP (NP \textbf{PN}) VP) & Subj. \\
	5.2 & (DNP  (NP \textbf{PN}) DEG) & Genitive \\
	6.6 & (IP (NP \textbf{PN}) VP PU) & Subj. \\
	38.0 & (IP (NP \textbf{PN}) (VP VV VP)) & Subj. \\
	56.0 & (IP (NP \textbf{PN}) (VP ADVP VP)) & Subj. \\
	77.0 & (IP ADVP (NP \textbf{PN}) VP) & Subj. \\
	81.0 & (IP (NP \textbf{PN}) (VP ADVP VP) PU) & Subj. \\
	81.0 & (IP (ADVP AD) (NP \textbf{PN}) VP) & Subj. \\
	93.5 & (PP P (NP \textbf{PN})) & Obj. of prep. \\
	93.5 & (IP (NP \textbf{PN}) (VP VV IP)) & Subj. \\
	93.6 & (VP VV (NP \textbf{PN}) IP) & Obj. of verb \\ \hline

\end{tabular}
}
\caption{Top subtree (depth=2) features involving pronouns (PN)}
\label{pronounFeat}
\end{table}

\begin{CJK}{UTF8}{gbsn}
The high ranking of pronoun-related features (4 out of the top 10 features involve pronouns) confirms the distinguishing power of pronoun usage. Crucially, it appears that pronouns in subject position or as a genitive (as part of DNP phrase such as \textit{他}的书, \textit{his book}), are more prominent than pronoun in the object position in translated texts. In fact, pronouns as the object of a preposition (captured by subtree ``(PP P (NP PN))'') ranked only about 93rd among all features. Also, pronouns as the object of a verb only shows up once in the top 100 features, and they are of the structure ``(VP VV (NP PN) IP)''. When searching for sentences with such structures (using Tregex), we almost always encounter phrases similar to ``make + pronoun + V.'', e.g. ``让 他们 懂得 ..." (\textit{make them understand ...}), where the pronoun is both the object of ``make'', and the subject of ``understand''. All this shows that the over-usage of pronouns in translated texts is more likely to occur in subject positions, or in a genitive complement, rather than as the direct object of a verb. Even when it appears in the object position, it appears to play both the roles of subject and object. To our knowledge, this characteristic has not been discussed in previous studies in translationese. 

If we examine the dependency features, we see the same pattern. Pronouns serving as the subject of verbs rank very high (5.4, 10, 17, 24, 35.6, see Table~\ref{tab:depfeats}), whereas pronouns as the object of verbs are not in the top 100 features (the highest ranking 191, VV\_DOBJ\_它 \textit{it}). Thus we see the two types of syntactic features (constituent trees and dependency trees) converging to the same conclusion. If we look at the pronoun issue from the opposite side, a reasonable consequence would be that in original texts, more common nouns should serve as the subject, which is indeed what we find. VV\_NSUBJ\_NN predicts ``original'' and ranks 94.8.

The conclusion concerning pronoun usage drawn from the ranking of syntactic features coincides with observation of (non-)pro-drop in English and Chinese. I.e., Chinese is pro-drop while Enlgish is not.
Thus, the overuse of pronouns in Chinese texts translated from English is an example of the interference effect \citep{toury1979interlanguage}, where translators are likely to carry over linguistic features in the source language to the target language. A further observation is that, in Chinese, subject pro-drop seems to be more frequent. The reason is that subject pro-drop does not require much context, while object-drop generally requires the dropped object to be discourse old
\citep[c.f.][]{li1981functional}.
This explains why pronoun overuse occurs more often in subject position in translated text, because object pro-drop in Chinese itself is less common in original Chinese text. 

We are not trying to imply that lexical features should not be used.  Rather, we want to stress that syntactic features offer a more in-depth and comprehensive picture to linguists interested in the style of translated text. The pronoun analysis presented above is only one such example. We can perform such analyses for any feature of interest and gain a deeper understanding of how they occur in both types of text.

\end{CJK}

\section{Conclusion and Future Work}
\begin{CJK}{UTF8}{gbsn}
To our knowledge, the current study is the first machine learning experiment on translated vs. original Chinese. We find that translationese can be identified with roughly the same high accuracy using either lexical \textit{n}-gram features or syntactic features. More importantly, we show how syntactic features can yield linguistically meaningful features that can help decipher differences in styles of translated and original texts. 
For example, translated Chinese features more determiners, subject-position pronouns, NP modifiers involving ``的'', and multiple NPs or VPs conjoined by the Chinese-specific punctuation ``、''.
Our methodology can, in principle, be applied to any stylistic comparisons in the digital humanities, and can yield stylistic insights much deeper than the pioneering work of \citet{federalistAuthorship1963}.

In future work, we will investigate tree substitution grammar (TSG), which extracts even deeper constituent trees \citep[c.f.][]{bayesianTSG2009}, and detailed feature interpretation for phrases headed by other tags (ADJP, PP, etc.) and for specific genres. It is also desirable to improve the accuracy of constituent parsers for Chinese, along the lines of \citep{lattice2013wang,parsingWang2014joint,lattice2017hu}, since accurate syntactic trees are the prerequisite for accurate feature interpretation. While the parser in this study works well, better parsers will undoubtedly be a plus.
\end{CJK}

\section*{Acknowledgement}
We thank Ruoze Huang, Jiahui Huang and Chien-Jer Charles Lin for helpful discussions, and the anonymous reviewers for their suggestions. Hai Hu is funded by China Scholarship Council. 

\bibliography{naaclhlt2018}
\bibliographystyle{acl_natbib}

\appendix

\end{document}